\documentclass[sigconf,nonacm,screen]{acmart}
\AtBeginDocument{%
  }

\usepackage{enumitem} 
\usepackage{subcaption} 
\usepackage{placeins}
\usepackage{float}  
\usepackage{algorithm}
\usepackage{algpseudocode}
\usepackage{multirow}  
\usepackage{booktabs}  
\usepackage{graphicx} 
\usepackage{array}
\usepackage{geometry}
\usepackage{fancyvrb}
\usepackage[utf8]{inputenc}
\usepackage{tcolorbox}
\usepackage{lipsum} 
\usepackage{listings}
\usepackage{xcolor} 
\usepackage{pifont}
\definecolor{mutedgreen}{RGB}{60,160,60}
\newcommand{\cmark}{\textcolor{mutedgreen}{\ding{51}}}
\newcommand{\xmark}{\textcolor{red}{\ding{55}}}
\AtBeginDocument{%
  \hypersetup{
    colorlinks=true,
    urlcolor=blue,
    citecolor=black,
    linkcolor=black
  }%
}
\begin{document}

\title{ToolForge: A Data Synthesis Pipeline for Multi-Hop Search without Real-World APIs}

\author{Hao Chen\textsuperscript{1,2}, Zhexin Hu\textsuperscript{2,3}, Jiajun Chai\textsuperscript{2}, Haocheng Yang\textsuperscript{2,4}, Hang He\textsuperscript{2,5}, Xiaohan Wang\textsuperscript{2}, \\Wei Lin\textsuperscript{2}, Luhang Wang\textsuperscript{1}, Guojun Yin\textsuperscript{2$\dagger$}, Zhuofeng Zhao\textsuperscript{1$\dagger$}} 
\affiliation{\institution{\textsuperscript{1}North China University of Technology, \textsuperscript{2}Meituan, \textsuperscript{3}Institute of Software, Chinese Academy of Sciences} \country{}} 
\affiliation{\institution{\textsuperscript{4}National University of Singapore, \textsuperscript{5}East China Normal University} \country{}}
\thanks{$\dagger$Corresponding author.}
\renewcommand{\shortauthors}{Meituan-AsX Team}
\begin{abstract}
Training LLMs to invoke tools and leverage retrieved information necessitates high-quality, diverse data. However, existing pipelines for synthetic data generation often rely on tens of thousands of real API calls to enhance generalization, incurring prohibitive costs while lacking multi-hop reasoning and self-reflection. To address these limitations, we introduce ToolForge, an automated synthesis framework that achieves strong real-world tool-calling performance by constructing only a small number of virtual tools, eliminating the need for real API calls. ToolForge leverages a (question, golden context, answer) triple to synthesize large-scale tool-learning data specifically designed for multi-hop search scenarios, further enriching the generated data through multi-hop reasoning and self-reflection mechanisms. To ensure data fidelity, we employ a Multi-Layer Validation Framework that integrates both rule-based and model-based assessments. Empirical results show that a model with only 8B parameters, when trained on our synthesized data, outperforms GPT-4o on multiple benchmarks. Our code and dataset are publicly available at \url{https://github.com/Buycar-arb/ToolForge}.
\end{abstract}





\maketitle

\begin{figure*}
    \centering
    \includegraphics[width=1.0\textwidth]{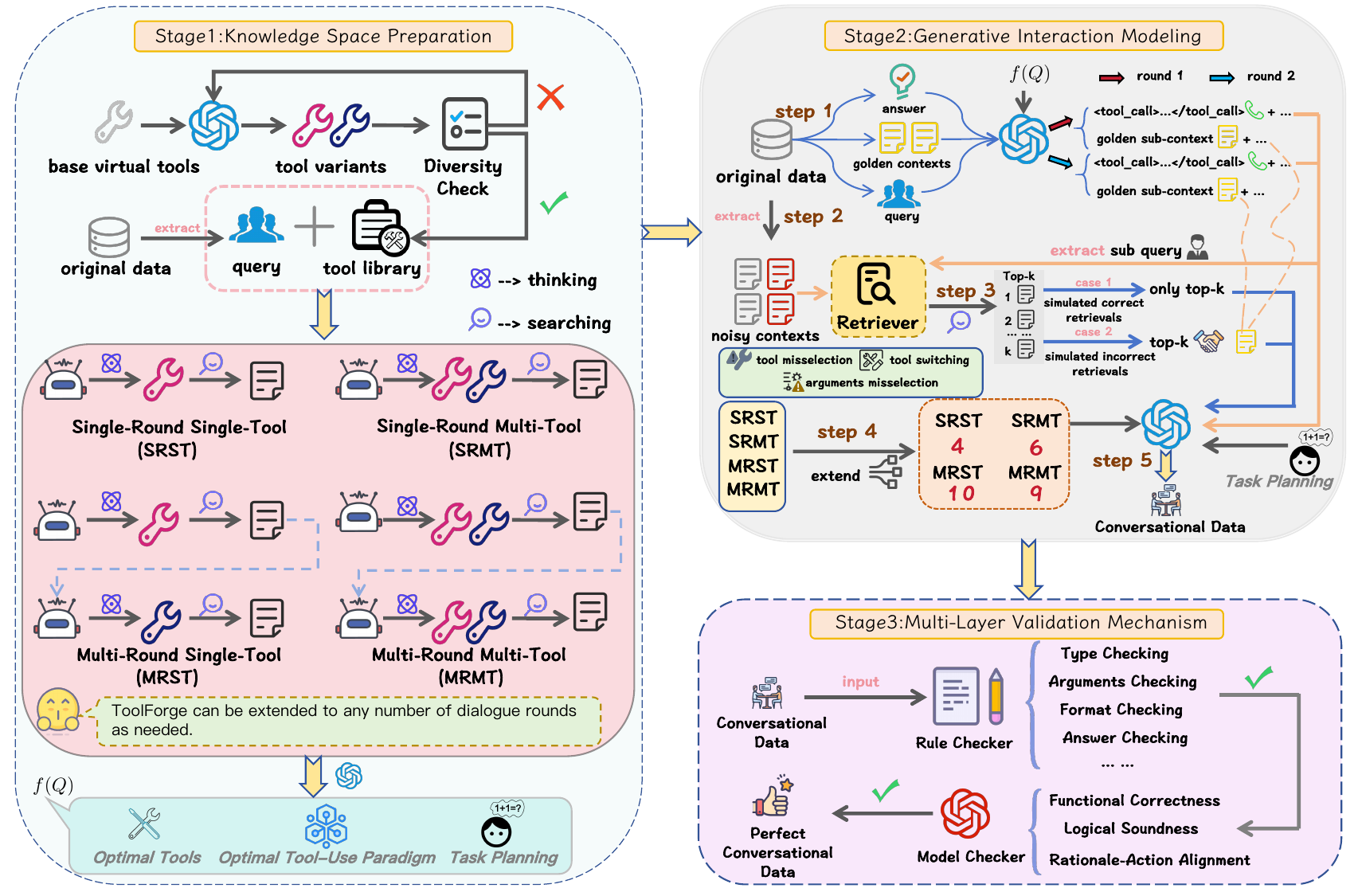}
    \Description{An overview diagram illustrating the three main modules of ToolForge: Knowledge Space Preparation (KSP), Generative Interaction Modeling (GIM), and Multi-Layer Validation (MLV).}
    \caption{The overall framework of ToolForge, which mainly consists of Knowledge Space Preparation (KSP), Generative Interaction Modeling (GIM), and Multi-Layer Validation (MLV).}
    \label{fig:main_picture}
\end{figure*}

\section{Introduction}
In recent years, large language models (LLMs) have demonstrated remarkable capabilities in natural language understanding \cite{yang2025qwen3, lin2025restreshapingtokenlevelpolicy}, particularly in search and tool learning. By integrating external tool APIs \cite{qu2025tool,chai2025rlfactory}, tool-augmented
LLMs have achieved a qualitative leap in practical applicability, enabling them to tackle complex real-world scenarios \cite{schick2023toolformer,qu2025tool}. Tool-calling mechanisms empower Large Language Models (LLMs) to interact with the external environment, enabling them to dynamically retrieve and access up-to-date information \citep{yuan2024easytool}. This strategy effectively mitigates the inherent limitations of static pretraining data and substantially expands their practical value across various application domains. Notable examples include workflow automation \cite{zhong2023llm4eda} and travel planning \cite{hao2024large}.

Training LLMs for tool-calling typically requires large-scale, high-quality synthetic data covering diverse scenarios \cite{liu2024toolace}. However, since training LLMs on a limited set of tools is insufficient for achieving robust generalization \cite{shi2025tool,mekala2024toolverifier}, existing data synthesis pipelines \cite{team2025kimi,liu2024toolace,wang2024appbench} commonly compensate for this deficiency by designing templates and executing tens of thousands of real-world API calls to obtain results \cite{qin2023toolllm} and construct training samples. To enhance the generalization capability of Large Language Models (LLMs), we utilize 19 virtual tools as surrogates for real-world API calls.

Real-world tasks often require \textbf{multi-hop reasoning} \cite{wang2022self}, i.e., a reasoning process that derives the final answer through multiple intermediate steps and logical chains \cite{yang2018hotpotqa,ho2020constructing}. Yet, most existing works concentrate on text-based multi-hop reasoning \cite{trivedi2022musique, press2022measuring}, lacking the capability to integrate with external tools. Concurrently, while research indicates that reflection enables models to inspect and revise their own reasoning processes \cite{shinn2023reflexion}, its potential in complex scenarios involving multi-hop reasoning and tool-calling remains underexplored. To bridge this gap, we devise four tool-calling paradigms and three error perturbation classes, thereby deriving 29 distinct interaction patterns that encompass complex, multi-turn tool-calling scenarios.

Ensuring the fidelity of complex, automatically synthesized data presents a significant challenge \cite{basu2024api,team2025kimi}. Existing validation approaches are often superficial, primarily verifying the syntactic correctness of tool-calling and the final answer consistency \cite{wang2024appbench,patil2024gorilla}. They largely overlook the semantic and logical integrity of intermediate reasoning steps, allowing subtle errors in multi-step chains to go undetected and thus compromising the overall data quality \cite{liu2024toolace}. To address this, we introduce a Multi-Layer Validation Framework that combines rule-based heuristics with model-based assessments, using Monte Carlo Tree Search (MCTS) \cite{silver2016mastering} for hard negative mining to substantially enhance validation robustness and coverage.

In summary, our key contributions are fourfold: 

\begin{itemize}[leftmargin=*]
  \item We introduce ToolForge, a novel automated synthesis framework that, given only a (question, golden context, answer) triple, can generate large-scale tool-calling data featuring multi-hop reasoning and self-reflection.
  \item We enhance the generalization capability of LLMs by leveraging virtual tools instead of real APIs, and incorporating reflection-driven multi-turn interactions to generate diverse reasoning-tool interaction patterns.
  \item ToolForge is inherently extensible, rather than being restricted to the 19 virtual tools and 29 interaction patterns instantiated in this paper. Additional virtual tools, new noise types, or more complex interaction motifs can be incorporated in a plug-and-play fashion without modifying the core pipeline.
  \item We demonstrate the state-of-the-art effectiveness of our approach. Our model, ToolForge-8B, developed by fine-tuning Qwen3-8B exclusively on our synthetic data, significantly outperforms strong proprietary models like GPT-4o across a range of challenging tool-calling benchmarks.
\end{itemize}

\section{Related works}
\textbf{\textit{Tool Learning.}} Tool learning has been shown to substantially enhance the performance of large language models (LLMs) on complex tasks \citep{qu2025tool,chai2025rlfactory}. By integrating external tools, LLMs are able to access real-time information \citep{chen2024advancing}, expand domain-specific knowledge \citep{m2024augmenting}, and process multimodal inputs \citep{wang2024genartist}. Approaches for training LLMs to invoke tools and leverage retrieved information can be broadly divided into two categories. The first is fine-tuning based methods \citep{liu2024toolace,team2025kimi,qin2023toolllm}, which rely on collecting large-scale input–output traces through tens of thousands of real API calls to train LLMs in tool usage. However, these methods suffer from prohibitive costs and limited scalability due to heavy reliance on real API calls. The second is non-fine-tuning methods \citep{yuan2024easytool,yao2023react}, which improve tool-calling ability through prompt optimization. While this reduces data construction cost, their capabilities are often restricted to relatively simple single-step reasoning, falling short in handling complex search scenarios that require multi-hop reasoning and multi-turn interactions \cite{tang2023toolalpaca}. Unlike these methods, ToolForge synthesizes complex tool-calling data with only a few virtual tools.

\noindent\textbf{\textit{Multihop Data.}} Multi-hop reasoning requires a system to integrate dispersed pieces of evidence into coherent reasoning chains \cite{yang2018hotpotqa}. Existing multi-hop datasets are mainly constructed through manual annotation \citep{ho2020constructing,trivedi2022musique} or template-based synthesis \citep{press2022measuring,talmor2018web,he2025localsearchbenchbenchmarkingagenticsearch}, both of which follow static paradigms. In addition, recent approaches leverage large language models to automatically synthesize multi-hop data \citep{liu2024toolace,team2025kimi,wang2024appbench}, aiming to reduce annotation costs and improve data diversity. However, these methods still rely on high-quality labeled trajectories, lack the ability to simulate dynamic tool interactions, and do not incorporate reflective processes. ToolForge addresses these limitations by shifting from static text-based reasoning to dynamic interactive reasoning.

\noindent\textbf{\textit{Data Synthesis.}} The scarcity of high-quality data remains a critical bottleneck for large language models (LLMs) \citep{bauer2024comprehensive}. Existing synthesis strategies include prompt templates \citep{wang2022self,xu2023wizardlm,yu2023metamath}, API-based generation \citep{basu2024api,liu2024toolace}, and dialogue-based construction \citep{chen2025acebench,patil2024gorilla,wang2024appbench}, but these methods are often complex to implement and incur high costs. To lower this barrier, ToolForge adopts a lightweight transformation strategy, ensuring both diversity and complexity while significantly reducing construction cost.

\section{Methods}
Prior studies \cite{mitra2024agentinstruct} have demonstrated that high-quality synthetic data plays a critical role in improving the capabilities of Tool-Augmented LLMs. Inspired by this observation, in this section we present the proposed ToolForge, whose framework is illustrated in Figure ~\ref{fig:main_picture}. The framework encompasses three primary modules: 1) \textbf{Knowledge Space Preparation}, 2) \textbf{Generative Interaction Modeling}, and 3) \textbf{Multi-Layer Validation}.
The overall success rate of ToolForge’s data synthesis process is shown in Table ~\ref{tab:success_rate}.

\subsection{Knowledge Space Preparation}
This module is the central component of our synthetic data generation pipeline, which transforms structured inputs into executable reasoning paths featuring tool-calling. The methodology is bifurcated into two primary stages. Section \ref{sec:Tool Construction and Diversification} constructing and diversifying a set of virtual tools to simulate real-world retrieval scenarios and Section \ref{sec:Tool-Calling Paradigm Design} extracting representative tool-calling paradigms from a base dataset comprising (question, golden context, answer) triples to cover a spectrum of real-world scenarios, from elementary to intricate.
\subsubsection{Tool Construction and Diversification}
\label{sec:Tool Construction and Diversification}
We first abstract retrieval domains from a large set of intent instances and design 19 domain specific virtual tools that span major areas, including but not limited to economics, politics, and science, forming a small set of base virtual tools. To enrich tool diversity without relying on massive real APIs, we then design a dual-gating mechanism \( \mathcal{J} \), illustrated as the \textbf{Diversity Check} in Figure~\ref{fig:main_picture}, to generate diverse tool variants by jointly applying semantic and textual similarity constraints, ultimately ensuring that the synthesized data exhibits sufficient diversity in tool-calling behaviors.
\begin{equation}
\mathcal{J}
= \mathbf{1}\!\left\{ \delta_{\mathcal S} \;\lor\; \text{Agg}_{\cos}(\tilde{T}, \mathcal{S}) > \theta_c \right\} 
\cdot 
\mathbf{1}\!\left\{ \text{Agg}_{\text{text}}(\tilde{T}, \mathcal{S}) < \theta_b \right\}
\end{equation}
\begin{align}
    \delta_{\mathcal S} &=
    \begin{cases}
    1, & \text{if } |\mathcal S| < 2, \\
    0, & \text{otherwise}.
    \end{cases} \label{eq:coldstart} \\[6pt]
    \text{Agg}_{\cos}(\tilde{T}, \mathcal S) &= \frac{1}{|\mathcal S|} \sum_{T_i \in \mathcal S} \text{Sim}_{\cos}(\tilde{T}, T_i), \\
    \text{Agg}_{\text{text}}(\tilde{T}, \mathcal S) &= \frac{1}{|\mathcal S|} \sum_{T_i \in \mathcal S} \hat{\text{Sim}}_{\text{BM25}}(\tilde{T}, T_i).
\end{align}
where \( \mathcal{S} \) denotes the set of already accepted tools, and \( \tilde{T} \) is a candidate variant tool. The threshold \( \theta_c \) is a lower bound on semantic similarity, and \( \theta_b \) is an upper bound on textual similarity. \( \text{Agg}_{\cos}(\tilde{T}, \mathcal{S})  \) and \( \text{Agg}_{\text{text}}(\tilde{T}, \mathcal{S})  \) denote the average semantic similarity and textual similarity between the candidate tool and the base tools, respectively. A candidate tool is incorporated into the final tool set \( \mathcal{S} \) only when it satisfies both semantic relevance and low textual redundancy. The logical OR operator \( \lor\ \) dictates that the semantic relevance check is automatically satisfied by the cold-start mechanism when \( |\mathcal{S}| < 2  \) to alleviate instability caused by small sample sizes; otherwise, it requires the aggregated cosine similarity to exceed its threshold.

\subsubsection{Tool-Calling Paradigm Design}
\label{sec:Tool-Calling Paradigm Design}
\begin{figure*}[t]
    \centering
    \includegraphics[width=1.0\textwidth]{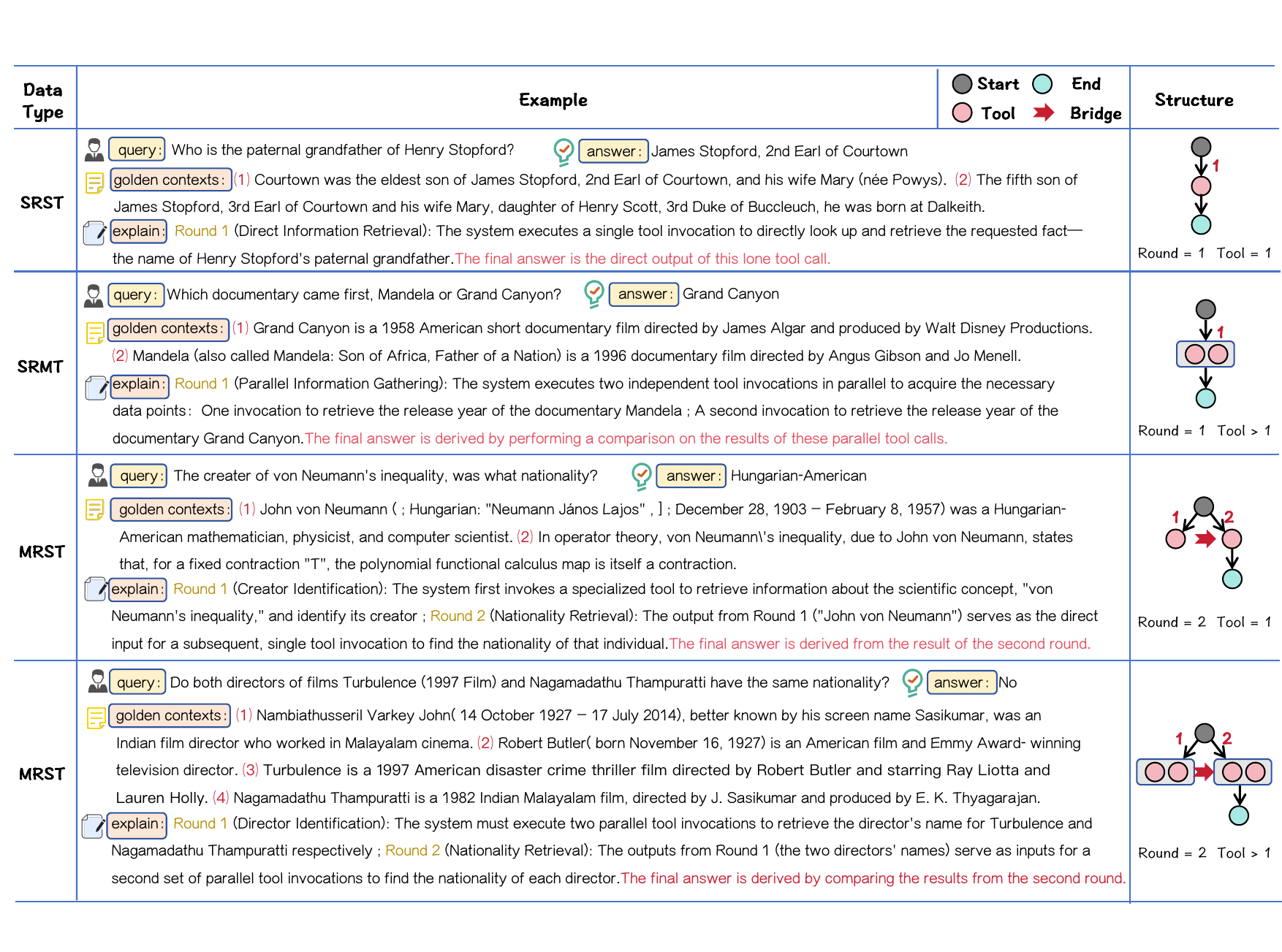}
    \caption{
        Illustration of the four tool-calling paradigms in ToolForge. While this figure illustrates the simplest form of each tool-calling paradigm for clarity, the full ToolForge dataset features instances with far more complex logic structures.
    }
    \label{fig:example_picture}
\end{figure*}

To systematically cover diverse retrieval scenarios and progressively enhance the reflective reasoning capabilities of the model, we categorize tool-calling behaviors into four paradigms, ranging from simple to complex, as illustrated in Figure \ref{fig:example_picture}. Each example in the figure shows a (question, golden context, answer) triple with the corresponding reasoning workflow on the right. Gray and cyan nodes indicate the start and end of reasoning, while pink nodes represent tool-calling. Red arrows denote cross-step dependencies where intermediate outputs feed into subsequent steps. \textbf{Round} and \textbf{Tool} specify the number of turns and tool-calling per turn.
\begin{itemize}[leftmargin=*]
  \item \textbf{\textit{Single-Round Single-Tool (SRST)}}. Involves the calling of a single tool within one turn to directly resolve a given problem. This represents the most fundamental tool-calling scenario.
  \item \textbf{\textit{Single-Round Multi-Tool (SRMT)}}. Requires the parallel or composite calling of multiple tools within a single turn. The complexity arises from the need to synthesize results and ensure synergy among the tools.
  \item \textbf{\textit{Multi-Round Single-Tool (MRST)}}. The reasoning process depends on the sequential execution of a single tool across multiple conversational turns. This paradigm emphasizes cross-turn contextual dependency and state maintenance.
  \item \textbf{\textit{Multi-Round Multi-Tool (MRMT)}}. Characterized by both cross-turn dependencies and the concurrent use of multiple tools. This represents the most sophisticated and complex mode of interaction.
\end{itemize}
Within these paradigms, the joint optimization of tool-calling sequences, tool-calling paradigms, and reasoning paths is formulated as a sequential decision process. We address this by optimizing these components sequentially according to a predefined priority. The procedure first selects the optimal tool-calling sequence 
\( \mathcal S^* \) from the set \( \mathcal S \). Conditioned on this sequence, we then determine the optimal tool-calling paradigm \( \mathcal C^* \). Finally, we deduce the corresponding optimal reasoning rationale \( \mathcal R^* \). This process is formalized as follows:
\begin{align}
\mathcal S^* &= \arg \max_{\mathcal S} P(\mathcal S \mid \mathcal Q), \\
\mathcal C^* &= \arg \max_\mathcal C P(\mathcal C \mid \mathcal Q, \mathcal S^*), \\
\mathcal R^* &= \arg \max_\mathcal R P(\mathcal R \mid \mathcal Q, \mathcal S^*, \mathcal C^*).
\end{align}
Therefore, the joint prediction function \( f(\mathcal Q) \) denotes the search for an optimal solution over the tool set \( \mathcal S \), the space of tool-calling paradigms \( \mathcal C \), and the space of reasoning rationales \( \mathcal R \), as shown in Equation~\ref{eq:joint_prediction}.
\begin{align}
f(\mathcal Q) = \arg\max_{\mathcal S,\, \mathcal C,\, \mathcal R} P(\mathcal S, \mathcal C, \mathcal R \mid \mathcal Q) 
= \langle \mathcal S^{*},\, \mathcal C^{*},\, \mathcal R^{*} \rangle
\label{eq:joint_prediction}
\end{align}
All outcomes were subjected to a rigorous cross-validation by several independent annotators to confirm their validity. Ablation studies of the Knowledge Space Preparation (KSP) framework are further discussed in Section \ref{sec:ksp_ablation}.

\subsection{Generative Interaction Modeling}
In this work, we propose a novel \textbf{Generative Interaction Modeling} (GIM) method to better simulate real-world tool-calling scenarios. The generation process is divided into two core stages: 1) Exploring Reasoning-Tool Interaction Patterns, 2) Input Preparation and Dialogue Generation.
\subsubsection{Exploring Reasoning-Tool Interaction Patterns}
\begin{figure}[t]
  \centering
  \includegraphics[width=\columnwidth]{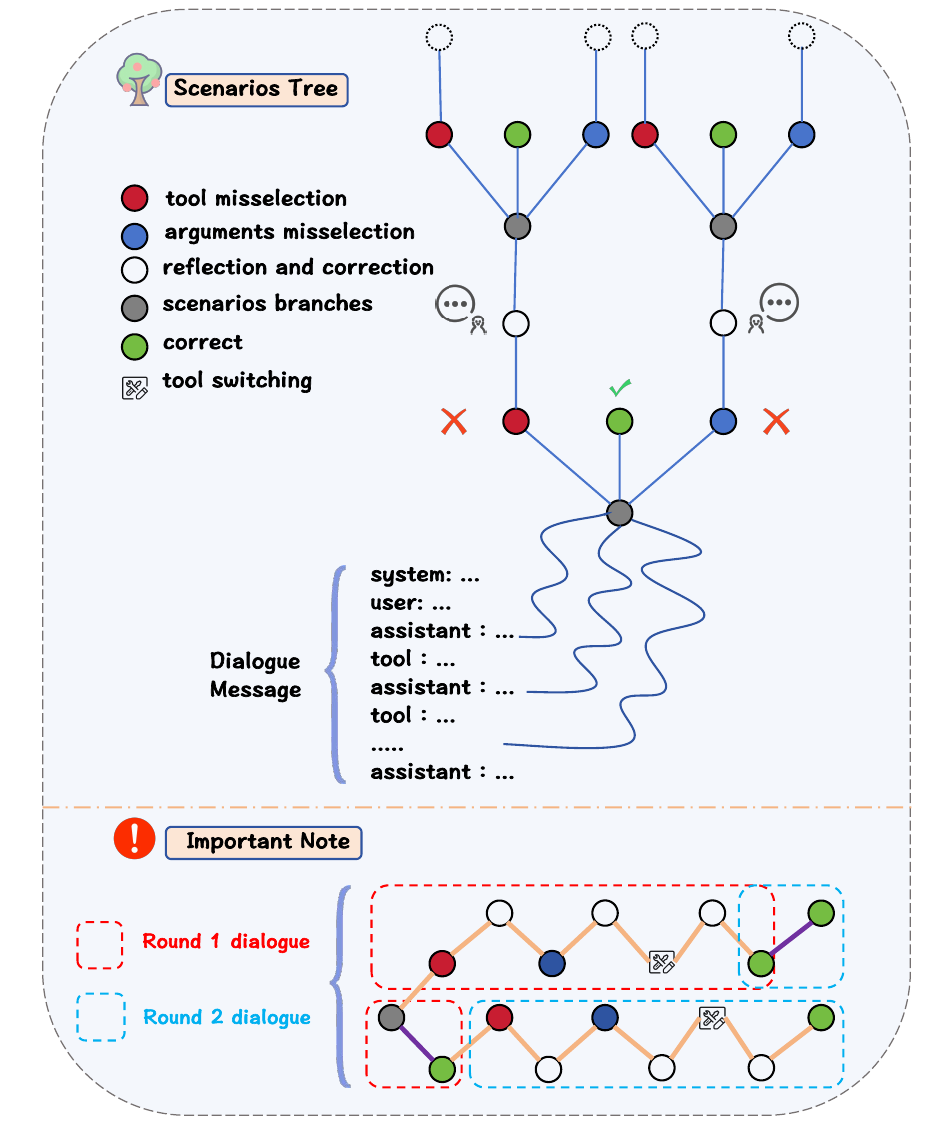}
  \caption{Scenario tree with three possible outcomes for an intermediate step within a single turn: (i) correct tool-calling, (ii) tool misselection, and (iii) argument misselection. The important note shows two tool-switching cases across a two-round dialogue.}
  \Description{A small scenario tree diagram illustrating three outcomes and a note with two tool-switching examples across two rounds.}
  \label{fig:tree_picture}
\end{figure}

To closely emulate the complex and diverse tool-calling scenarios in the real world, we start from four representative tool-calling paradigms. We then introduce three classes of error perturbations: tool misselection, arguments misselection, and tool switching. Through the combination and extension of these paradigms and perturbations, we synthesize 29 representative reasoning-tool interaction patterns \( \mathcal{M} = \{ \mathrm{flow}_i \}_{i=1}^{29} \), as illustrated in the Figure~\ref{fig:tree_picture}.
\subsubsection{Input Preparation and Dialogue Generation}
ToolForge requires no manually annotated golden context. Instead, GIM leverages the supporting evidence already provided in existing QA datasets and automatically retrieves any missing passages when necessary, thus enabling fully automatic synthesis without additional annotation costs. We then construct the structured inputs for dialogue generation through a multi-phase process detailed in Algorithm \ref{alg:dialogue_generation}. 
The process begins with a \textbf{Planning Phase}, where a large language model ($\mathcal{L}$) is prompted with the question ($\mathcal{Q}$), answer ($\mathcal{A}$), and reasoning rationale ($\mathcal{R}^*$) to generate an execution trace ($\tau$). This trace specifies the necessary sequence of tool- calling ($\text{calls}_1$) and their corresponding golden sub-contexts ($\text{refs}_1$).
Subsequently, the \textbf{Augmentation Phase} prepares two distinct information sets for each planned action. We retrieve candidate documents ($K_i$) from a corpus ($\mathcal{D}$) via the BM25 algorithm. These retrieved-only documents constitute the incorrect information ($I_i^{\text{bad}} = K_i$), while the correct information ($I_i^{\text{good}}$) is formed by augmenting them with the golden sub-context ($K_i \cup \text{ref}_i$). This pairing is crucial for training the model to handle both noisy and ideal contexts.
In the \textbf{Generation Phase}, a generator model is conditioned on the plan and the prepared information pairs ($\mathcal{I}_1$) to synthesize the multi-turn dialogue. Finally, an \textbf{Assembly Phase} parses this output and structures it into the final data format ($\mathcal{D}_{\text{final}}$), ensuring the creation of high-quality data ready for validation.

\begin{algorithm}[t]
\caption{Synthetic Dialogue Generation}
\label{alg:dialogue_generation}
\begin{algorithmic}[1]
\Require 
    $\mathcal Q, \mathcal A, \mathcal R^*, \mathcal G^*, \mathcal{T}, \mathcal D, \mathcal{L}$
\Ensure 
    Final dialogue data $\mathcal{D}_{\text{final}}$

\Statex \textbf{$\blacktriangleright$ Phase 1: Planning}
\State $\mathcal{P}_{\text{plan}} \gets \text{ConstructPrompt}(\mathcal Q, \mathcal A, \mathcal R^*, \dots)$
\State $\tau \gets \mathcal{L}(\mathcal{P}_{\text{plan}})$ \Comment{Generate execution trace}
\State $(\text{calls}_1, \text{refs}_1) \gets \Call{ParseTrace}{\tau}$

\Statex \textbf{$\blacktriangleright$ Phase 2: Augmentation}
\State $\mathcal{I}_1 \gets []$ \Comment{Initialize list for information pairs}
\For{each $(\text{call}_i, \text{ref}_i)$ in $(\text{calls}_1, \text{refs}_1)$}
    \State $\text{query} \gets \Call{ExtractQuery}{\text{call}_i}$
    \State $K_i \gets \Call{BM25}{\mathcal D, \text{query}}$ \Comment{Retrieve candidates}
    \State $I^{\text{good}}_i \gets K_i \cup \text{ref}_i$ \Comment{Augmented information}
    \State $I^{\text{bad}}_i \gets K_i$ \Comment{Retrieved-only information}
    \State Append $(I^{\text{good}}_i, I^{\text{bad}}_i)$ to $\mathcal{I}_1$
\EndFor

\Statex \textbf{$\blacktriangleright$ Phase 3: Generation}
\State $\mathcal{P}_{\text{dialogue}} \gets \text{ConstructPrompt}(\mathcal Q, \text{calls}_1, \mathcal{I}_1, \dots)$
\State $\mathcal{D} \gets \mathcal{L}(\mathcal{P}_{\text{dialogue}})$ \Comment{Generate multi-turn dialogue}
\State $\mathcal{D}_{\text{parsed}} \gets \Call{ParseDialogue}{\mathcal{D}}$

\Statex \textbf{$\blacktriangleright$ Phase 4: Assembly}
\State $\mathcal{D}_{\text{final}} \gets \Call{AssembleOutput}{\mathcal{D}_{\text{parsed}}, \mathcal{I}_1, \dots}$
\State \Return $\mathcal{D}_{\text{final}}$

\end{algorithmic}
\end{algorithm}

\subsection{Multi-Layer Validation}
The quality of synthetic data is paramount for a model's tool-calling proficiency, as structural flaws or logical inconsistencies can derail its reasoning processes and degrade overall performance \cite{liu2024apigen}. We therefore propose a \textbf{Multi-Layer Validation} (MLV) framework to verify our synthesized data, as illustrated in Figure ~\ref{fig:main_picture}. It is composed of a \textbf{Rule Verification Layer} and a \textbf{Model Verification Layer}, with all results being overseen by human experts.

\subsubsection{Rule Verification Layer}
This layer employs an automated rule checker, which is comprised of a meticulously designed set of nine static rules covering four key dimensions: consistency in dialogue format and structure, correctness of tool protocols, overall dialog correctness, and the traceability of data samples. Its objective is twofold: 
\begin{itemize}[leftmargin=*]
\item Ensure that the synthesized data adheres to predefined formatting specifications.
\item Identify any unsubstantiated elaborations or fabricated information by systematically comparing the model's input against its output, thus preventing factually deviant outcomes caused by model hallucination.
\end{itemize}

For instance, when validating the correctness of a tool protocol, our rule checker sequentially extracts the content within the assistant's \texttt{<tool\_call>} tags, comparing parameter changes field-by-field to ensure adherence to the predefined tool protocol. These procedures allow us to validate the correc tness and executability of tool-calling without the need for actual execution, which enhances efficiency and reduces deployment overhead.

\begin{table*}[t]
\centering
\caption{Main experimental results. The best performance is set in bold. $^\dagger$/$^\star$ represents in-domain/out-domain datasets.}
\label{tab:retrieval}
\setlength\tabcolsep{0.9pt} 
\renewcommand{\arraystretch}{1.1} 
\small 
\begin{tabular}{l*{20}{c}} 
\toprule
\multirow{2}{*}{\bf Model} 
  & \multicolumn{2}{c}{\bf NQ$^\star$} 
  & \multicolumn{2}{c}{\bf PopQA$^\star$} 
  & \multicolumn{2}{c}{\bf TriviaQA$^\star$} 
  & \multicolumn{2}{c}{\bf SQuAD$^\star$} 
  & \multicolumn{2}{c}{\bf Bamboogle$^\star$} 
  & \multicolumn{2}{c}{\bf MusiQue$^\star$} 
  & \multicolumn{2}{c}{\bf SRST$^\dagger$}
  & \multicolumn{2}{c}{\bf SRMT$^\dagger$}
  & \multicolumn{2}{c}{\bf MRST$^\dagger$}
  & \multicolumn{2}{c}{\bf MRMT$^\dagger$} \\
\cmidrule(lr){2-3} \cmidrule(lr){4-5} \cmidrule(lr){6-7} \cmidrule(lr){8-9} \cmidrule(lr){10-11} \cmidrule(lr){12-13} \cmidrule(lr){14-15} \cmidrule(lr){16-17} \cmidrule(lr){18-19} \cmidrule(lr){20-21}
& EM & F1 & EM & F1 & EM & F1 & EM & F1 & EM & F1 & EM & F1 & EM & F1 & EM & F1 & EM & F1 & EM & F1 \\

\midrule
\multicolumn{21}{l}{\textit{\textbf{Wikipedia Retriever + Basic Search}}}\\
GPT-4o-2024-08-06 & 19.53 & 27.83 & 20.25 & 25.79 & 13.30 & 30.88 & 16.60 & 25.59 & 59.20 & 70.69 & 12.60 & 16.52 & \underline{\textbf{62.00}} & 71.39 & 55.00 & 61.54 & 46.00 & 59.74 & 57.00 & 59.06 \\
Qwen3-4B & 14.07 & 21.05 & 20.26 & 25.86 & 8.95 & 22.56 & 10.17 & 17.85 & 24.00 & 30.44 & 7.60 & 10.84 & 4.50 & 16.36 & 21.50 & 25.43 & 13.00 & 18.76 & 13.00 & 13.54  \\
Qwen3-8B & 13.55 & 19.18 & 16.66 & 20.52 & 8.01 & 18.27 & 8.74 & 14.43 & 24.80 & 32.20 & 22.40 & 25.67 & 6.00 & 13.82 & 50.00 & 53.28 & 12.50 & 16.24 & 50.50 & 50.50 \\
Qwen3-32B & 23.63 & 33.83 & 29.32 & 37.83 & 15.36 & 34.48 & 18.03 & 28.51 & 41.60 & 50.87 & 31.80 & 38.35 & 22.50 & 37.49 & 66.50 & 71.63 & 32.50 & 40.46 & 40.00 & 40.00 \\

\midrule
\multicolumn{21}{l}{\textit{\textbf{Wikipedia Retriever + Function Call}}}\\
GPT-4o-2024-08-06 & 16.68 & 23.27 & 23.65 & 29.17 & 12.20 & 25.15 & 19.60 & 30.07 & 56.00 & 69.02 & 35.00 & 44.92 & 32.00 & 37.30 & 34.00 & 38.79 & 25.50 & 30.34 & 30.00 & 31.25 \\
Qwen3-4B & 19.39 & 28.72 & 23.73 & 30.61 & 12.87 & 30.22 & 13.40 & 22.50 & 39.20 & 51.14 & 7.80 & 18.05 & 21.00 & 33.72 & 7.50 & 11.19 & 26.00 & 36.54 &6.00 & 12.11  \\
Qwen3-8B & 20.80 & 30.17 & 24.21 & 31.61 & 14.04 & 32.38 & 14.19 & 22.84 & 41.60 & 54.15 & 19.60 & 28.34 & 23.00 & 37.33 & 22.00 & 29.34 & 30.00 & 38.98 & 26.00 & 28.60 \\
Qwen3-32B & 16.34 & 26.47 & 23.16 & 30.91 & 13.91 & 32.28 & 13.79 & 22.96 & 40.80 & 54.69 & 23.40 & 32.51 & 32.00 & 46.34 & 18.50 & 26.48 & 27.50 & 36.58 & 37.50 & 41.40 \\
Qwen3-235B & 30.20 & 40.82 & 34.20 & 40.80 & \underline{\textbf{19.60}} & \underline{\textbf{41.87}} & 23.80 & \underline{\textbf{36.78}} & \underline{\textbf{59.60}} & \underline{\textbf{71.15}} & \underline{\textbf{45.20}} & \underline{\textbf{53.89}} & 58.50 & \underline{\textbf{72.45}} & 71.00 & 71.12 & 44.50 & 57.69 & 73.50 & 74.23 \\
\textbf{ToolForge-8B} & \underline{\textbf{30.25}} & \underline{\textbf{40.92}} & \underline{\textbf{35.20}} & \underline{\textbf{41.08}} & 16.80 & 36.98 & \underline{\textbf{24.00}} & 32.75 & 48.00 & 58.66 & 37.80 & 45.10 & 20.00 & 32.99 & \underline{\textbf{71.00}} & \underline{\textbf{74.41}} & \underline{\textbf{48.00}} & \underline{\textbf{61.55}} & \underline{\textbf{77.50}} & \underline{\textbf{77.63}} \\


\bottomrule
\end{tabular}
\end{table*}

\begin{table*}[t]
\centering
\caption{Overall performance of ToolForge-8B on eight downstream reasoning benchmarks.}
\label{tab:overall_8tasks}
\setlength\tabcolsep{5pt}
\renewcommand{\arraystretch}{1.1}
\small
\begin{tabular}{l*{8}{c}}
\toprule
\multirow{2}{*}{\bf Model} 
  & \multicolumn{1}{c}{\bf MATH-500} 
  & \multicolumn{1}{c}{\bf AIME-2025} 
  & \multicolumn{1}{c}{\bf MMLU-Redux} 
  & \multicolumn{1}{c}{\bf GPQA-Diamond} 
  & \multicolumn{1}{c}{\bf IFEval} 
  & \multicolumn{1}{c}{\bf Arena-Hard} 
  & \multicolumn{1}{c}{\bf HumanEval} 
  & \multicolumn{1}{c}{\bf MBPP} \\
\cmidrule(lr){2-9}
& Acc & Pass@64 & Acc & Pass@10 & Acc & Win-rate & Pass@1 & Pass@1 \\
\midrule
Qwen3-8B (Base)      &  \underline{\textbf{97.4}}  &  67.3  &  87.5  &  \underline{\textbf{62.0}}  &  85.0  &  85.8  &  57.9  &  72.1  \\
ToolForge-8B (Ours)  &  96.3  &  \underline{\textbf{71.2}}  &  \underline{\textbf{88.4}}  &  61.3  &  \underline{\textbf{86.8}}  &  \underline{\textbf{87.1}}  &  \underline{\textbf{58.3}}  &  \underline{\textbf{73.4}}  \\
\bottomrule
\end{tabular}
\end{table*}

\subsubsection{Model Verification Layer}
 While the rule-based layer excels at enforcing structural and procedural integrity, it has a limited capacity for evaluating semantic coherence and the plausibility of the reasoning process. A model might generate an output that is formally impeccable yet logically flawed or misaligned with the task objective. Consequently, we introduce the Model Verification Layer to assess sample quality from a higher-level semantic dimension. This assessment is structured around three core principles:
 \begin{itemize}[leftmargin=*]
  \item \textbf{\textit{Correctness of Tool-Calling}}. Evaluates whether the retrieval tool is appropriately selected and query parameters are well-configured.
  \item \textbf{\textit{Logical Soundness of Reasoning}}. Assesses whether the intermediate reasoning steps and the analysis of retrieved content are logically coherent and free from factual errors.
  \item \textbf{\textit{Consistency between Thought and Action}}. Verifies that the reasoning chain serves as a rational basis for the actions taken (i.e., tool-calling) and the final conclusion.
\end{itemize}
Each aspect is evaluated by an individual expert agent, powered by an LLM.

To systematically evaluate the performance of our MLV framework, we designed and constructed a comprehensive, three-tiered evaluation benchmark, comprising the following: (1) Within 29 reasoning–tool interaction patterns, we randomly sampled 20 instances per pattern. All samples were annotated by human experts, resulting in 580 positive examples in total. (2) For each of the nine static rules, we first introduced a set of common real-world errors. We then employed Monte Carlo Tree Search \cite{silver2016mastering} to explore diverse combinations of these errors. Finally, for each rule, we sampled the top-5 most challenging negative instances across the 29 interaction patterns. (3) For each of the 29 interaction patterns, we generated three distinct samples, each specifically designed to violate one of the three core semantic principles. Evaluation results on this benchmark are presented in Table~\ref{tab:mlv_metrics}. Ablation studies of the MLV framework are further discussed in Section \ref{sec:mlv_ablation}. 

\begin{table*}[t]
\centering
\begin{minipage}{0.48\linewidth}
\centering
\caption{Success rate of the four synthesis routes after the Generate \& MLV stages.}
\label{tab:success_rate}
\begin{tabular}{lc}
\toprule
\textbf{Route} & \textbf{Success Rate (\%)} \\
\midrule
SRST & 91.70 \\
SRMT & 89.50 \\
MRST & 86.30 \\
MRMT & 83.80 \\
\bottomrule
\end{tabular}
\end{minipage}
\hfill
\begin{minipage}{0.48\linewidth}
\centering
\caption{Evaluation metrics used to assess the accuracy of the MLV module.}
\label{tab:mlv_metrics}
\begin{tabular}{lc}
\toprule
\textbf{Metric} & \textbf{Score (\%)} \\
\midrule
Accuracy & 98.47 \\
Precision & 96.55 \\
Recall & 98.25 \\
F1 Score & 97.39 \\
\bottomrule
\end{tabular}
\end{minipage}
\end{table*}


\section{Experiment}
In this section, we conduct extensive experiments on a diverse set of 10 real-world benchmarks to validate the effectiveness of our synthetic data.

\subsection{Experiment Setup}
\subsubsection{Datasets}
We evaluate ToolForge on seven benchmarks and four newly proposed benchmarks, categorized as follows: (1) General Question Answering: NQ \cite{kwiatkowski2019natural}, TriviaQA \cite{joshi2017triviaqa}, SQuAD \cite{rajpurkar2016squad}, PopQA \cite{mallen2022not}, and SRST; (2) Multi-Hop Question Answering: MusiQue \cite{trivedi2022musique}, Bamboogle \cite{press2022measuring}, SRMT, MRST, and MRMT. These benchmarks encompass a diverse range of search-with-reasoning challenges, enabling a comprehensive evaluation of ToolForge.
\subsubsection{Baselines.}
We benchmark our method against two categories of state-of-the-art baselines:
(i) the proprietary, API-based model GPT-4o \cite{hurst2024gpt}; and (ii) leading open-source function-calling models, exemplified by the Qwen 3 family. 

\begin{table*}[t]
\centering
\caption{Ablation results showing the impact of synthetic data ratios on the performance of ToolForge-8B. The best performance is set in bold. $^\dagger$/$^\star$ represents in-domain/out-domain datasets.}
\label{tab:rate}
\setlength\tabcolsep{3pt} 
\renewcommand{\arraystretch}{1.1} 
\small 
\begin{tabular}{l*{12}{c}} 
\toprule
\multirow{2}{*}{\bf Model} 
  & \multicolumn{2}{c}{\bf NQ$^\star$} 
  & \multicolumn{2}{c}{\bf Bamboogle$^\star$} 
  & \multicolumn{2}{c}{\bf SRST$^\dagger$}
  & \multicolumn{2}{c}{\bf SRMT$^\dagger$}
  & \multicolumn{2}{c}{\bf MRST$^\dagger$}
  & \multicolumn{2}{c}{\bf MRMT$^\dagger$} \\
\cmidrule(lr){2-3} \cmidrule(lr){4-5} \cmidrule(lr){6-7} \cmidrule(lr){8-9} \cmidrule(lr){10-11} \cmidrule(lr){12-13} 
& EM & F1 & EM & F1 & EM & F1 & EM & F1 & EM & F1 & EM & F1 \\

\midrule
\multicolumn{13}{l}{\textit{\textbf{Wikipedia Retriever + Function Call}}}\\
ToolForge-8B (9:1) & 30.25 & 40.92 & 48.00 & 58.66 & 20.50 & 34.87 & 73.00 & 75.91 & \underline{\textbf{48.00}} & \underline{\textbf{61.55}} & \underline{\textbf{77.50}} & \underline{\textbf{77.63}} \\
ToolForge-8B (8:2) & 29.58 & 40.23 & 44.80 & 55.01 & 20.50 & 33.69 & \underline{\textbf{73.00}} & \underline{\textbf{76.03}} & 42.00 & 55.01 & 75.50 & 75.50 \\
ToolForge-8B (7:3) & 29.89 & 39.94 & 40.00 & 51.13 & 19.00 & 32.74 & 60.50 & 64.96 & 43.00 & 49.92 & 57.50 & 57.54 \\
ToolForge-8B (6:4) & 29.86 & 40.10 & 44.80 & 55.85 & 22.00 & 35.95 & 71.50 & 75.11 & 44.50 & 52.17 & 66.50 & 66.50 \\
ToolForge-8B (5:5) & 29.56 & 40.66 & 48.00 & 58.59 & 18.50 & 31.76 & 67.00 & 70.45 & 43.50 & 50.98 & 56.00 & 56.00 \\
ToolForge-8B (4:6) & 28.95 & 39.52 & 48.00 & 59.25 & \underline{\textbf{23.00}} & \underline{\textbf{36.46}} & 70.50 & 72.82 & 44.50 & 51.82 & 52.00 & 52.03 \\
ToolForge-8B (3:7) & \underline{\textbf{30.58}} & \underline{\textbf{41.02}} & \underline{\textbf{49.60}} & \underline{\textbf{60.01}} & 19.00 & 32.39 & 68.00 & 76.66 & 47.00 & 53.66 & 52.50 & 52.03 \\
ToolForge-8B (2:8) & 29.92 & 40.65 & 42.40 & 52.94 & 19.50 & 33.98 & 63.50 & 67.41 & 47.50 & 54.83 & 47.00 & 47.00 \\
ToolForge-8B (1:9) & 29.72 & 39.54 & 45.60 & 56.57 & 22.50 & 34.72 & 70.50 & 73.34 & 48.00 & 53.99 & 48.50 & 48.50 \\

\bottomrule
\end{tabular}
\end{table*}

\subsubsection{Evaluation Metrics.}
Evaluation is conducted on the test or
validation sets of 10 benchmarks to assess both in-domain and out-of-domain performance. Consistent with prior work \cite{jin2025search}, we report model performance using the Exact Match (EM) and F1 metrics under deterministic decoding (temperature = 0). We adopt two interaction modes with Wikipedia in our experiments: Basic Search and Function Calling. In the \textit{\textbf{Basic Search}} mode, the model directly issues natural language queries to retrieve relevant passages, using the format \texttt{<search>query</search>}. In contrast, the \textit{\textbf{Function Call}} mode follows a structured tool-calling paradigm, where the model interacts with Wikipedia through predefined functions using the format:
\texttt{<tool\_call>tool\_name + tool\_arguments
</tool\_call>}. This structured interface enables more controlled and explicit retrieval compared to direct search.
In addition, we adopt a retrieval-based evaluation setting, where the model is provided with the top-10 passages retrieved from Wikipedia. This setting reflects a realistic open-domain scenario and enables us to evaluate the model’s ability to reason over retrieved contexts.
Details regarding the choice of the retrieval top-k are discussed in the Appendix ~\ref{app:topk_select}. 

\subsubsection{Implementation Details.} For data synthesis, we use HotpotQA \cite{yang2018hotpotqa} and 2WikiMultiHopQA \cite{ho2020constructing} as the original data.
We manually design 19 base virtual tools and construct 20 diverse variants per tool, thereby improving the diversity and robustness of the synthesized tool-calling trajectories.
Finally, we apply ToolForge’s three-stage pipeline to generate 4,250 training samples.

For training, we employ full-parameter supervised fine-tuning (SFT) to train the open-source LLM, Qwen3-8B \cite{yang2025qwen3}. The training process begins with the official Qwen3-8B model and finetunes on a dataset of 4,250 samples, preserving the original distribution with a 9:1 ratio, where SRST accounts for 89.4\%, and SRMT, MRST, and MRMT each constitute 3.5\%. We run the full training in \texttt{ms-Swift} \cite{zhao2025swift} and adopt a standard chat template to ensure compatibility between the training data format and Qwen3's input format. Detailed training parameters are provided in Appendix~\ref{app:training_config}. We further analyze the effect of different single-hop and multi-hop data ratios in Section \ref{sec:data_ratios}.

During inference, we adopt a deterministic decoding strategy with temperature = 0. To ensure fair comparison across all methods in Table~\ref{tab:retrieval}, we use the same retriever, the same number of retrieved documents, the same knowledge corpus, and the same prompt, with the full prompt template provided in Appendix~\ref{app:prompt}.

\begin{figure}[t]
    \centering
    \includegraphics[width=\columnwidth]{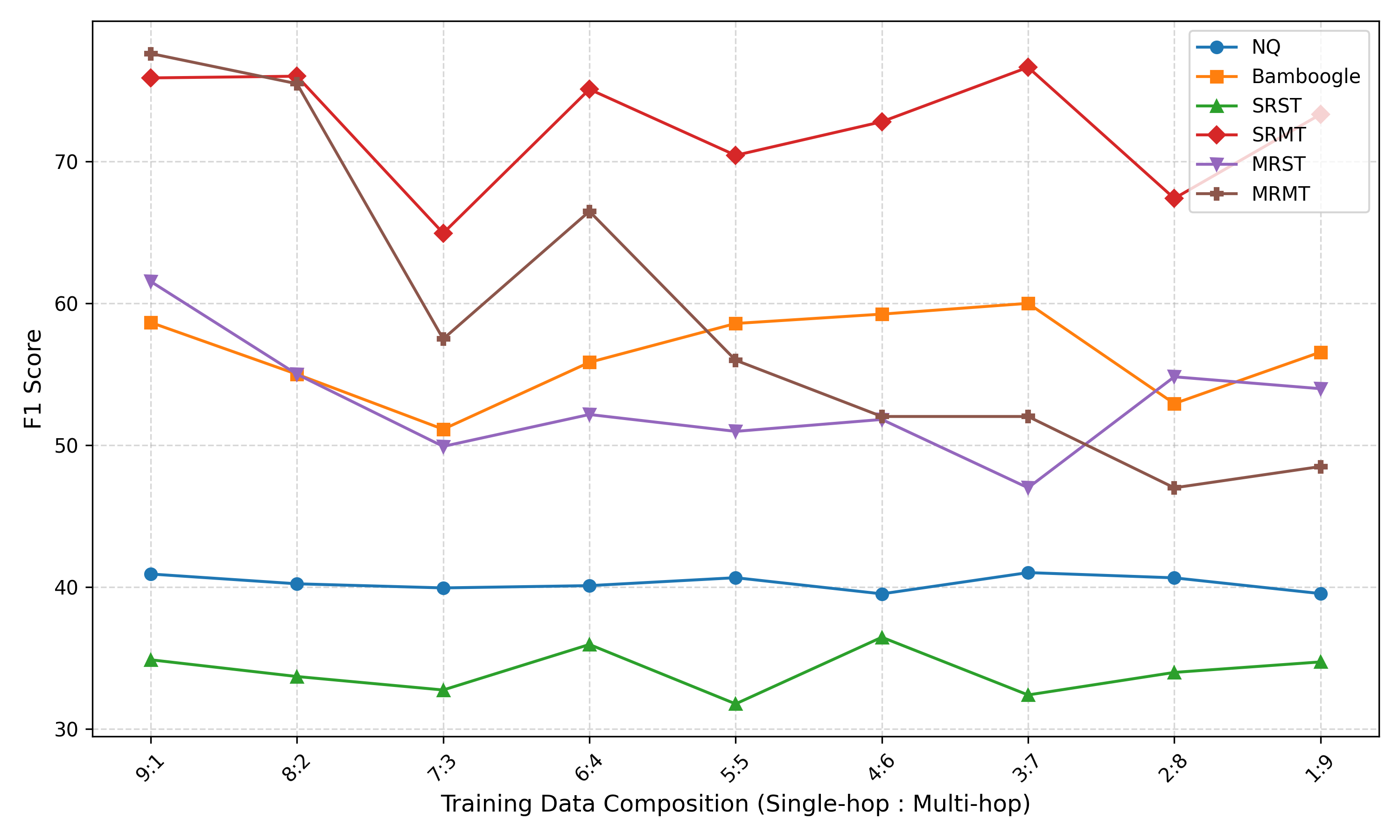}
    \caption{
        Effect of Single-hop and Multi-hop Data Ratios.
    }
    \label{fig:data_ratio_ablation_all_in_one}
\end{figure}
\subsection{Main Results}

As shown in Table \ref{tab:retrieval}, ToolForge-8B, which is specifically trained on our synthetic dataset, demonstrates state-of-the-art overall performance across a wide range of question-answering benchmarks. In the Function Call setting, our model proves superior to both leading proprietary and open-source models.

We observe that when transitioning from Basic Search to Function Call retrieval, models such as GPT-4o and the Qwen3 series suffer significant performance degradation across multiple benchmarks. This phenomenon highlights the distinct challenges inherent in the tool-calling task. In contrast, ToolForge-8B effectively mitigates this performance gap. ToolForge-8B surpasses the powerful GPT-4o model on 8 out of 10 benchmarks and outperforms the much larger Qwen3-32B model on 9 out of 10 benchmarks. 

Notably, despite having significantly fewer parameters, Tool-Forge-8B demonstrates performance comparable to the much larger Qwen3-235B, surpassing its Exact Match (EM) scores on 6 of the benchmarks. This underscores the efficiency and effectiveness of our method. Overall, ToolForge-8B achieves the best comprehensive performance, excelling on both single-hop QA tasks and complex multi-hop reasoning challenges. These results validate the versatility and advanced capabilities of our model across diverse multi-hop search scenarios.


    
    

    

\begin{table}[t]
\centering
\caption{Ablation study results.}
\label{tab:ablation-results}
\small
\setlength\tabcolsep{3pt}  
\renewcommand{\arraystretch}{1.0}  
\begin{tabular}{@{}lcccccccc@{}}  
\toprule
\multirow{2}{*}{\textbf{Model}} & \multicolumn{2}{c}{\textbf{SRST}} & \multicolumn{2}{c}{\textbf{SRMT}} & \multicolumn{2}{c}{\textbf{MRST}} & \multicolumn{2}{c}{\textbf{MRMT}} \\
\cmidrule(lr){2-3} \cmidrule(lr){4-5} \cmidrule(lr){6-7} \cmidrule(lr){8-9}
& \textbf{EM} & \textbf{F1} & \textbf{EM} & \textbf{F1} & \textbf{EM} & \textbf{F1} & \textbf{EM} & \textbf{F1} \\
\midrule
\multicolumn{9}{l}{\textit{\textbf{BM25 + Function Call}}} \\
GPT-4o-2024-08-06 & 65.5 & 77.6 & 68.0 & 71.4 & 62.5 & 72.7 & 89.0 & 89.0\\
Claude-Sonnet-4 & 40.5 & 55.3 & 63.2 & 66.7 & 49.0 & 60.3 & 82.1 & 80.6 \\
Qwen3-4B & 45.5 & 54.9 & 8.5 & 15.9 & 38.0 & 49.2 & 14.0 & 17.4  \\
Qwen3-8B & 53.0 & 61.5 & 25.0 & 28.3 & 51.5 & 59.6 & 33.0 & 33.5  \\
Qwen3-32B & 61.0 & 70.6 & 22.5 & 30.0 & 50.0 & 60.1 & 66.5 & 67.2 \\
Qwen3-235B & 77.5 & 87.8 & 83.5 & 88.6 & 74.1 & 80.3 & 96.5 & 96.5 \\
\midrule
\textbf{ToolForge-8B} & \underline{\textbf{78.3}} & \underline{\textbf{88.2}} & \underline{\textbf{88.5}} & \underline{\textbf{91.4}} & \underline{\textbf{75.0}} & \underline{\textbf{81.8}} & \underline{\textbf{99.5}} & \underline{\textbf{99.5}} \\
\bottomrule
\end{tabular}
\end{table}
\subsection{Zero-Shot Task Generalization}
To evaluate the generalization of ToolForge-8B beyond retrieval, we assess its zero-shot performance on four unseen tool-calling benchmarks and eight downstream reasoning tasks.
As a primary evaluation step, we conducted experiments across four domains: mathematical reasoning (MATH-500 \cite{lightman2023let} and AIME-2025 \cite{AIME2025}), general reasoning (MMLU-Redux \cite{gema2025we} and GPQA-Diamond \cite{rein2024gpqa}), alignment evaluation (IFEval \cite{zhou2023instruction} and Arena-Hard \cite{li2024crowdsourced}), and code generation (HumanEval \cite{chen2021evaluating} and MBPP \cite{austin2021program}). As shown in Table~\ref{tab:overall_8tasks}, ToolForge-8B outperforms the pre-fine-tuning Qwen3-8B in six out of eight tasks, demonstrating its strong generalization capability.

Furthermore, as indicated in Table~\ref{tab:tooluse}, ToolForge-8B achieves significant improvements on several unseen tool-calling benchmarks, surpassing both Qwen3-8B and the larger Qwen3-32B models on BFCL-v3 \cite{patil2025bfcl}, $\tau$-bench \cite{yao2024tau}, and ToolBench \cite{qin2023toolllm}, while maintaining competitive performance on NFCL \cite{nexusraven}. These results demonstrate that ToolForge-8B does not compromise its downstream reasoning performance, while also exhibiting strong generalization to zero-shot tool-calling tasks.

\begin{table}[t]
\centering
\caption{Tool-calling benchmark results across four datasets.}
\label{tab:tooluse}
\small
\setlength\tabcolsep{3pt}  
\renewcommand{\arraystretch}{1.0}

\begin{tabular}{@{}lcccc@{}}
\toprule
\multirow{2}{*}{\textbf{Model}} 
  & \multicolumn{1}{c}{\textbf{BFCL-v3}} 
  & \multicolumn{1}{c}{\boldmath$\tau$\textbf{-bench}} 
  & \multicolumn{1}{c}{\textbf{NFCL}} 
  & \multicolumn{1}{c}{\textbf{ToolBench}} \\
\cmidrule(lr){2-2} \cmidrule(lr){3-3} \cmidrule(lr){4-4} \cmidrule(lr){5-5}
& \textbf{Acc} & \textbf{Acc} & \textbf{Acc} & \textbf{Acc} \\
\midrule

Qwen3-32B & 16.6 & 31.3 & \underline{\textbf{50.6}} & 49.7 \\
Qwen3-8B  & 13.9 & 17.5 & 41.6 & 36.8 \\
\textbf{ToolForge-8B} 
    & \underline{\textbf{27.3}} 
    & \underline{\textbf{36.4}} 
    & 43.2 
    & \underline{\textbf{62.4}} \\

\bottomrule
\end{tabular}
\end{table}

\subsection{Ablation Study}
\subsubsection{Effect of Single-hop and Multi-hop Data Ratios}
\label{sec:data_ratios}
We compare the performance of LLMs trained on synthetic data with different single-hop and multi-hop data ratios across six benchmarks, as shown in Table \ref{tab:rate}. A total of 4,250 samples are partitioned between single-hop (SRST) and multi-hop (SRMT, MRST, MRMT) instances according to a predefined ratio, with the multi-hop allocation being uniformly distributed among the three subsets.

As shown in Figure~\ref{fig:data_ratio_ablation_all_in_one}, we make the following key observations: Single-hop dominant ratios yield the best overall performance, excessive multi-hop data leads to performance degradation. ToolForge-8B trained with a 9:1 ratio of single-hop to multi-hop data achieves the highest overall performance across benchmarks, including both in-domain and out-of-domain benchmarks. This indicates that a single-hop dominant composition, with a small proportion of multi-hop data, is most effective for training under limited data budgets.
Notably, even a small amount of multi-hop data is sufficient to teach the model basic multi-hop reasoning patterns, enabling it to generalize well across tasks.

\subsubsection{Effect of Retrieval Quality}
\label{sec:Effect of Retrieval Quality}
To mitigate the interference of retrieval failures in model evaluation, we replaced the standard Wikipedia corpus with a controlled, smaller-scale corpus constructed from golden and noisy contexts, ensuring more precise coverage. This setup allows for a more faithful assessment of the model's core reasoning capabilities. The results in Table \ref{tab:ablation-results} show that ToolForge-8B consistently outperforms strong baselines such as GPT-4o, Claude-Sonnet-4, and Qwen3 series across multiple public benchmarks. Its performance is particularly sensitive to retrieval quality: switching from the expansive Wikipedia corpus to our controlled corpus resulted in average relative improvements of 31.2\% in Exact Match (EM) and 28.5\% in F1 score. This highlights that ToolForge-8B can effectively leverage high-quality retrieval contexts to significantly enhance its multi-hop reasoning abilities and overall performance.
\begin{table*}[t]
\centering
\caption{Ablation study of the MLV framework, evaluating the impact of removing the Rule Verification Layer (RVL) and the Model Verification Layer (MVL).}
\label{tab:mlv_ablation}
\setlength\tabcolsep{0.92pt} 
\renewcommand{\arraystretch}{1.1} 
\small 
\begin{tabular}{l*{20}{c}} 
\toprule
\multirow{2}{*}{\bf Model} 
  & \multicolumn{2}{c}{\bf NQ$^\star$} 
  & \multicolumn{2}{c}{\bf PopQA$^\star$} 
  & \multicolumn{2}{c}{\bf TriviaQA$^\star$} 
  & \multicolumn{2}{c}{\bf SQuAD$^\star$} 
  & \multicolumn{2}{c}{\bf Bamboogle$^\star$} 
  & \multicolumn{2}{c}{\bf MusiQue$^\star$} 
  & \multicolumn{2}{c}{\bf SRST$^\dagger$}
  & \multicolumn{2}{c}{\bf SRMT$^\dagger$}
  & \multicolumn{2}{c}{\bf MRST$^\dagger$}
  & \multicolumn{2}{c}{\bf MRMT$^\dagger$} \\
\cmidrule(lr){2-3} \cmidrule(lr){4-5} \cmidrule(lr){6-7} \cmidrule(lr){8-9} \cmidrule(lr){10-11} \cmidrule(lr){12-13} \cmidrule(lr){14-15} \cmidrule(lr){16-17} \cmidrule(lr){18-19} \cmidrule(lr){20-21}
& EM & F1 & EM & F1 & EM & F1 & EM & F1 & EM & F1 & EM & F1 & EM & F1 & EM & F1 & EM & F1 & EM & F1 \\
\midrule
\multicolumn{21}{l}{\textit{\textbf{Wikipedia Retriever + Function Call}}}\\
Only RVL & 3.80 & 5.51 & 5.00 & 6.42 & 2.40 & 5.24 & 1.60 & 3.74 & 5.60 & 7.10 & 2.60 & 3.23 & 1.00 & 6.08 & 11.50 & 12.68 & 1.00 & 2.34 & 8.54 & 8.54 \\
Only MVL & 1.80 & 2.87 & 4.00 & 4.85 & 0.40 & 2.57 & 1.60 & 2.71 & 1.60 & 2.09 & 4.00 & 4.62 & 0.00 & 6.77 & 8.00 & 10.05 & 1.50 & 1.83 & 9.50 & 9.65  \\

RVL + MVL & 30.25 & 40.92 & 35.20 & 41.08 & 16.80 & 36.98 & 24.00 & 32.75 & 48.00 & 58.66 & 37.80 & 45.10 & 20.00 & 32.99 & 71.00 & 74.41 & 48.00 & 61.55 & 77.50 & 77.63 \\


\bottomrule
\end{tabular}
\end{table*}

\subsubsection{Ablation Study on KSP}
\label{sec:ksp_ablation}
\begin{figure}[t]
  \centering
  \begin{subfigure}[t]{0.49\columnwidth}
    \centering
    \includegraphics[width=\linewidth]{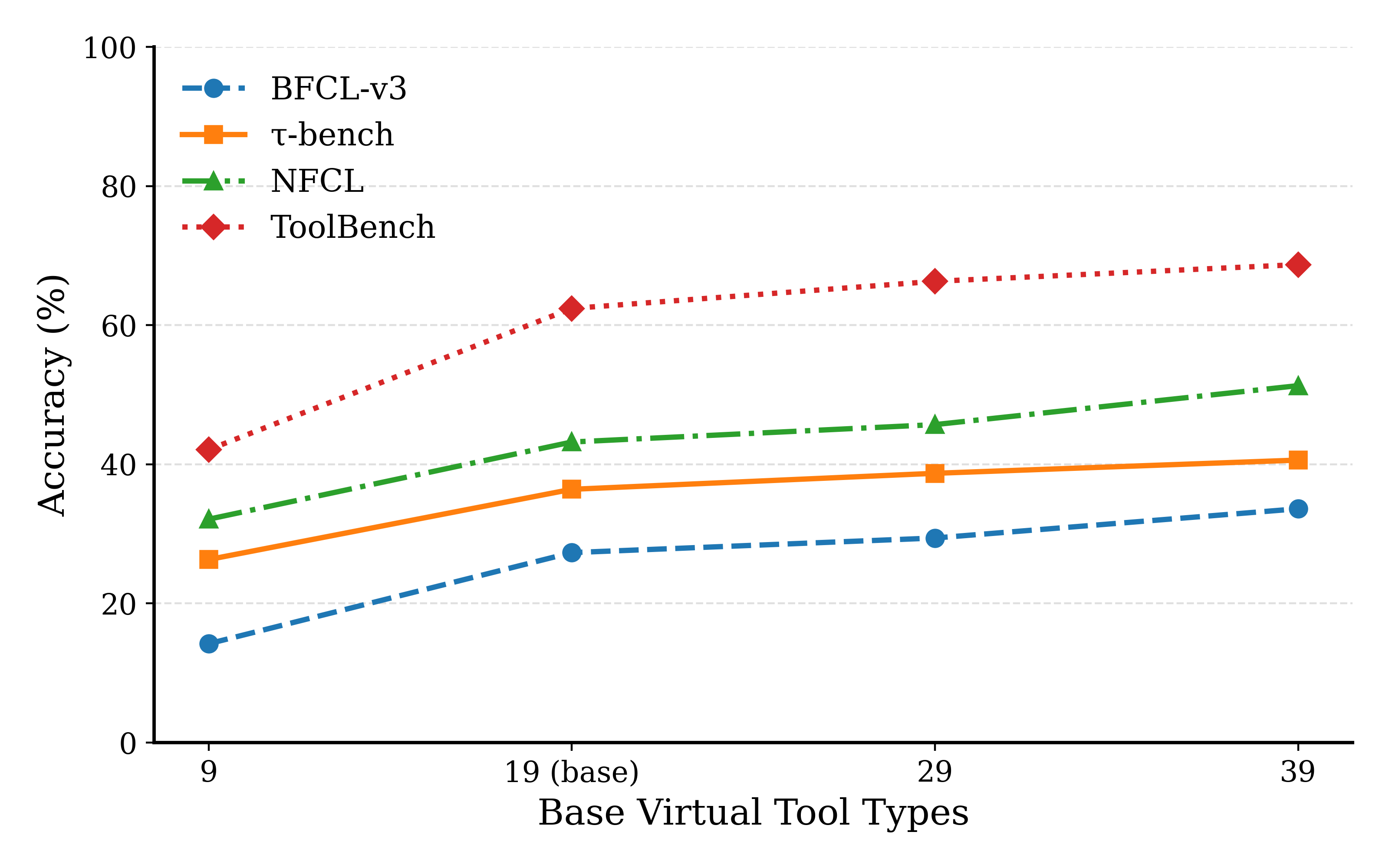}
    \caption{}
    \label{fig:tool_image_left}
  \end{subfigure}
  \hfill
  \begin{subfigure}[t]{0.49\columnwidth}
    \centering
    \includegraphics[width=\linewidth]{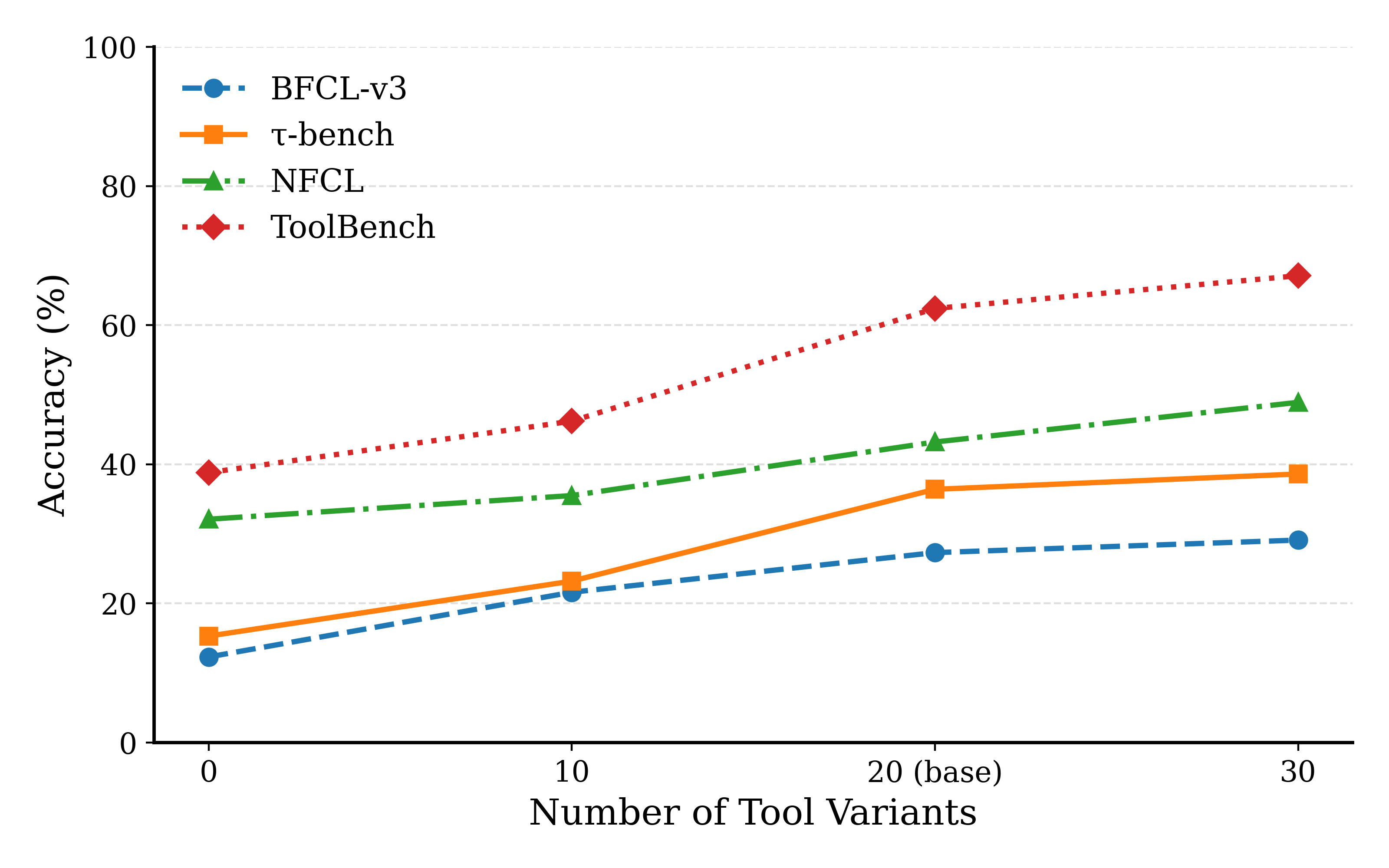}
    \caption{}
    \label{fig:tool_image_right}
  \end{subfigure}

  \caption{Experimental results of ablations on virtual tool design.}
  \label{fig:tool_image}
\end{figure}

We conduct two ablation studies to systematically analyze the role of tool diversity in our data synthesis pipeline: (i) We vary the number of virtual tool types while fixing 20 variants per tool, where each setting includes a general-purpose tool to ensure universal coverage across tool-calling scenarios; the results are shown in \ref{fig:tool_image_left}. (ii) We fix the number of tool types to 19 and vary the number of variants generated for each tool; the results are shown in \ref{fig:tool_image_right}.

In both experiments, “base” denotes the default ToolForge-8B configuration, which uses 19 virtual tools and 20 variants per tool.
The results demonstrate that increasing both the number of virtual tool types and the number of behavioral variants consistently improves performance across all benchmarks. In particular, expanding tool coverage yields substantial gains, while increasing intra-tool behavioral diversity leads to consistent additional improvements, underscoring the effectiveness of the KSP module in structuring tool taxonomy and variant diversity during data synthesis.


\subsubsection{Ablation Study on MLV}
\label{sec:mlv_ablation}
As shown in Table \ref{tab:success_rate}, skipping MLV produces 8.3\%–16.2\% noisy samples, including formatting errors and logical inconsistencies. With a small SFT dataset (only 4,250 instances), Table \ref{tab:mlv_ablation} shows that such noisy supervision significantly degrades model quality. Although the Model Verification Layer can theoretically cover most of the cases handled by the Rule Verification Layer, relying solely on model-based validation greatly increases computation cost. The Rule Verification Layer efficiently filters low-level and structural errors, while the Model Verification Layer focuses on the more challenging semantic inconsistencies. Therefore, both layers are necessary in practice. These results further highlight the critical role of the MLV in judging the quality of synthesized data during training data construction.

\section{Conclusion}
In this paper, we introduce ToolForge, a scalable framework for synthesizing large-scale, tool-augmented reasoning datasets with multi-hop and reflection capabilities. ToolForge leverages a small set of virtual tools and a lightweight data transformation pipeline to efficiently generate diverse and complex reasoning–tool interaction patterns, eliminating the need for real API calls. To ensure data quality, we propose a Multi-Layer Validation (MLV) framework that combines rule-based and model-based assessments, further enhanced by active hard negative mining. Extensive experiments demonstrate that ToolForge-8B, trained solely on our synthetic data, achieves strong performance across both in-domain and out-of-domain benchmarks, while maintaining robust generalization to zero-shot tool-calling and downstream reasoning tasks.Overall, ToolForge offers a practical and extensible solution for building high-quality synthetic training pipelines for tool-augmented LLMs.We hope this work encourages further exploration of scalable, verification-aware data synthesis for training reliable tool-augmented language models.


\bibliographystyle{ACM-Reference-Format}
\bibliography{sample-base}


\appendix
\section{Appendix}
\subsection{Training Configuration}
\label{app:training_config}
We adopt a full-parameter tuning strategy, training for two epochs on four NVIDIA A100 80GB GPUs using DeepSpeed ZeRO-2 for memory-efficient distributed training. Training utilizes \texttt{bfloat16} precision and Flash Attention 2 to accelerate computation and reduce memory usage. The learning rate is set to \(1 \times 10^{-5}\) with a constant schedule and a 5\% warmup phase. We use a per-device batch size of 2 with gradient accumulation steps of 1, resulting in an effective batch size of 8. The maximum sequence length is 12,000 tokens. We optimize with AdamW (weight decay 0.01). Training completes in approximately 2 hours. 
\subsection{Ablation on Retrieval Top-k Selection}
\label{app:topk_select}
\begin{figure}[t]
  \centering

  \begin{subfigure}[t]{0.48\columnwidth}
    \centering
    \includegraphics[width=\linewidth]{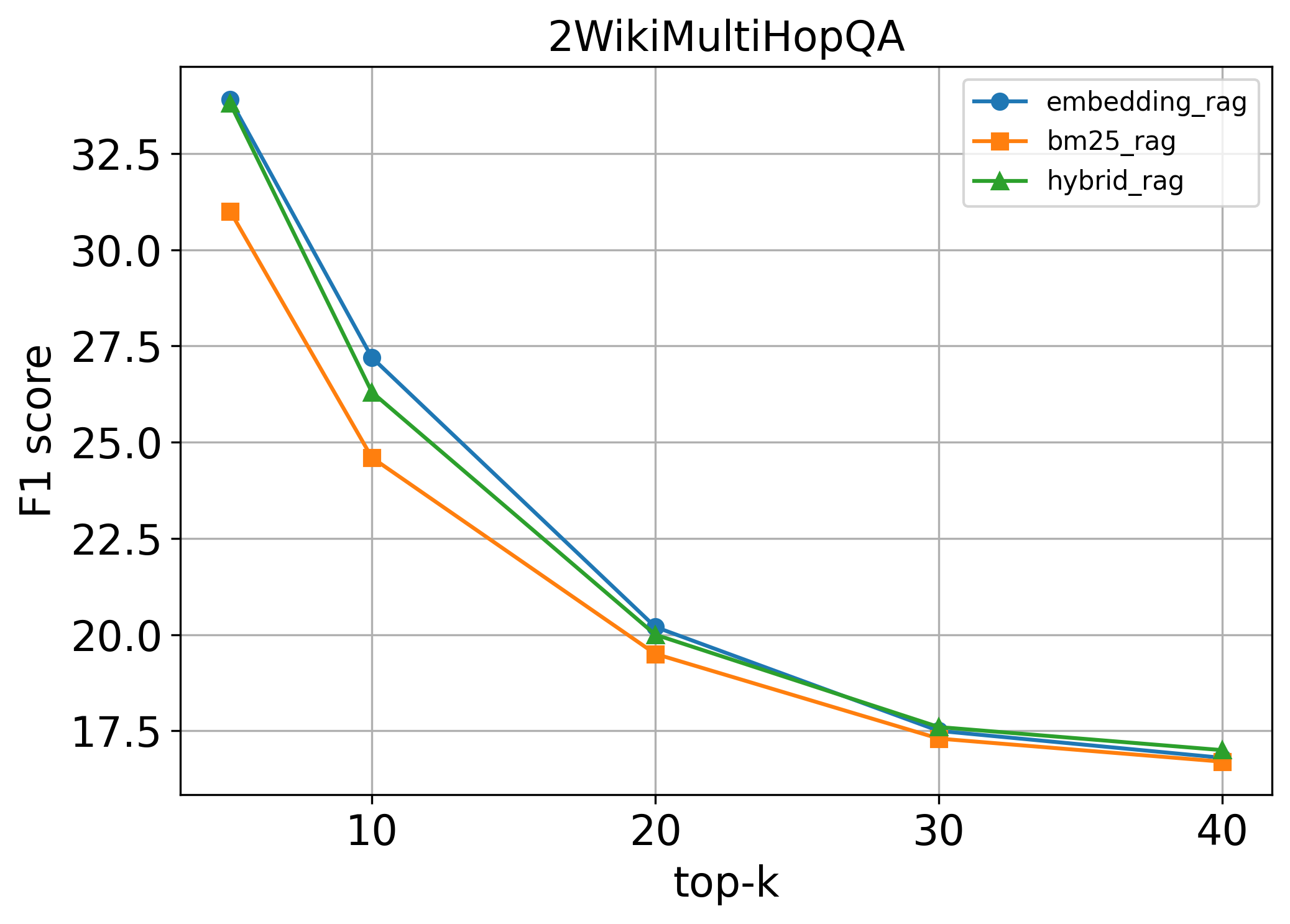}
    \caption{}
    \label{fig:subfig_a}
  \end{subfigure}
  \hfill
  \begin{subfigure}[t]{0.48\columnwidth}
    \centering
    \includegraphics[width=\linewidth]{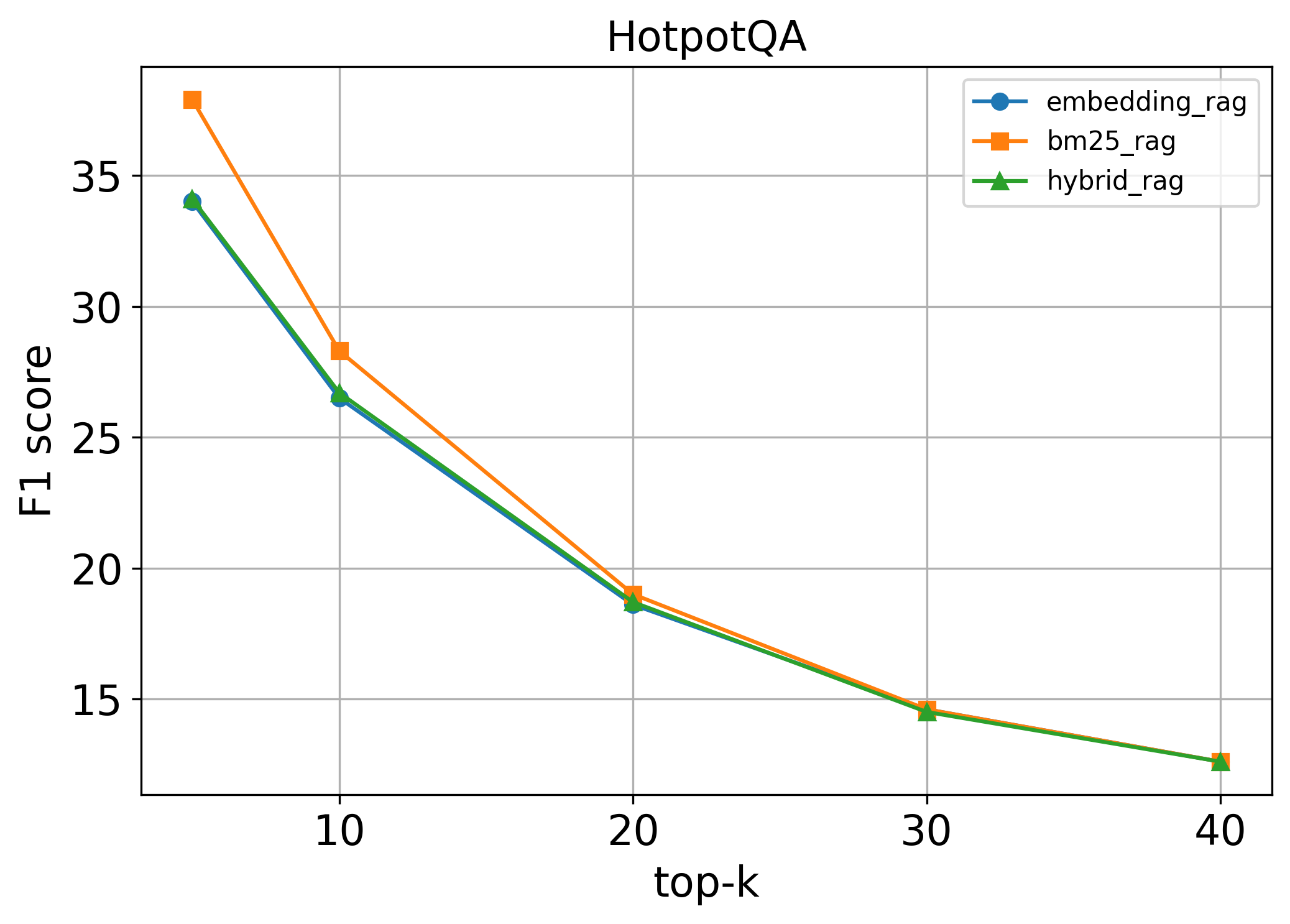}
    \caption{}
    \label{fig:subfig_b}
  \end{subfigure}

  \vspace{0.4em} 

  \begin{subfigure}[t]{0.48\columnwidth}
    \centering
    \includegraphics[width=\linewidth]{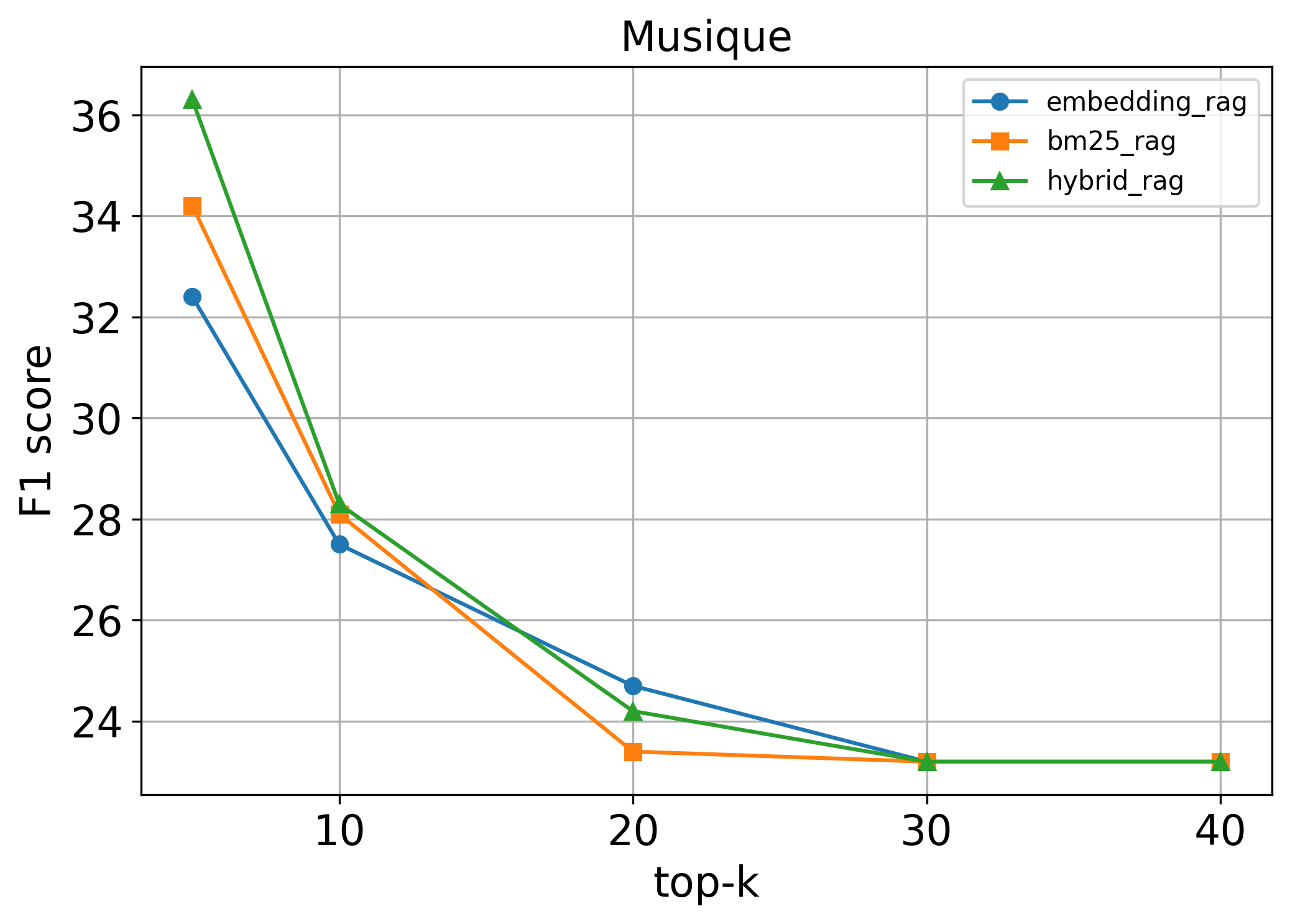}
    \caption{}
    \label{fig:subfig_c}
  \end{subfigure}

  \caption{Performance of different retrieval methods under varying top-$k$ context sizes.}
  \label{fig:three_subfigures}
\end{figure}

We conducted systematic ablation studies on various retrieval methods and top-k configurations. As illustrated in Figure \ref{fig:three_subfigures}, smaller top-k values generally yielded higher F1 scores, whereas larger values introduced noise and degraded performance. Balancing efficiency and efficacy, we ultimately selected BM25 with a top-10 setting as our default retrieval configuration.

\subsection{Case Study}
\label{app:case_study}
To demonstrate ToolForge's advantages in real-world scenarios, we present a case study comparing ToolForge-8B with GPT-4o on a complex multi-hop question requiring tool correction and reflective reasoning. This question is from the PopQA dataset: \textbf{Question:} "Who was the director of The Band?".
Table~\ref{tab:case_study} illustrates that the comparative analysis of the responses generated by the two models reveals significant differences in their reasoning strategies and information retrieval capabilities. ToolForge-8B exhibits superior performance in addressing ambiguous queries through a systematic approach involving multi-step reasoning and self-reflective optimization, whereas GPT-4o’s single-step methodology results in incorrect attribution.

ToolForge-8B’s architecture explicitly integrates a three-phase interaction flow—planning, analysis, and conclusion—with mandatory reflection checkpoints. When the initial search for “The Band director information” yielded generic musical leadership roles instead of specific film credits, the model identified the ambiguity and reformulated the query to “The Band film director information.” This iterative refinement, facilitated by the culture\_arts\_sports\_search tool’s structured parameters (work\_identifiers, artist\_or\_creator\_identifiers, categories), enabled precise identification of the 1978 Israeli film “The Band”. The subsequent search accurately retrieved the critical information that Avi Nesher directed this film, with the model performing a final validation to ensure consistency between the answer and the retrieved evidence.

In contrast, GPT-4o’s approach lacks structured iteration. Although it correctly recognized the cultural context and employed the same search function, its query—“The Band director”—without refined parameters, led to an incorrect documentary attribution to Daniel Roher. The model failed to distinguish between different works sharing the name “The Band” and did not implement secondary validation steps when initial results suggested ambiguity. This limitation underscores the absence of explicit mechanisms for resolving conflicting information or dynamically adjusting search strategies based on intermediate results.

The technical strength of ToolForge-8B lies in its deliberate decomposition of complex queries into manageable sub-problems. By first establishing that “The Band” refers to a film (rather than a music group or generic band concept) through contextual parameterization, and subsequently targeting director information using category filters, the model constructs a logical reasoning chain. 

\onecolumn
\begin{table*}[t]
\centering
\caption{Comparison of reasoning processes between GPT-4o and ToolForge-8B}
\label{tab:case_study}
\begin{tabular}{>{\bfseries}p{0.45\textwidth} >{\bfseries}p{0.55\textwidth}}
\toprule
GPT-4o & ToolForge-8B \\
\midrule

<think>

1. Identify the problem type: The question is about identifying the director of a film or documentary titled "The Band." This falls under the category of culture, arts, and sports.

2. Analyze the tool result (if applicable): Not applicable yet, as no tool has been called.

3. Perform a reflection check: Ensure that the search is directed towards film or documentary information, specifically focusing on the director's role. Consider if "The Band" might refer to a specific film or documentary about a musical group.

I will use the culture\_arts\_sports\_search function to find information about the director of "The Band."

</think>

<culture\_arts\_sports\_search>

query: "The Band director"

</culture\_arts\_sports\_search> 

\par\noindent\rule{\linewidth}{0.2pt}

<think>

1. Identify the problem type: The search results should provide information about the director of "The Band," which is likely a film or documentary.

2. Analyze the tool result: The search results indicate that "The Band" is a documentary directed by Daniel Roher. This documentary focuses on the musical group known as The Band.

3. Perform a reflection check: Ensure that the information aligns with the context of the question, confirming that Daniel Roher is indeed the director of the documentary about The Band.

The search results have provided the necessary information to answer the question accurately.

</think>

<answer>Daniel Roher</answer>

&
<think>

This question asks who the director of "The Band" is. I need to first understand what "The Band" refers to. It could be a film, a music group, or another type of work. Let me search for information related to "The Band" to determine its nature and director.

</think>

<tool\_call>

{"name": "culture\_arts\_sports\_search", "arguments": {"query": "The Band director information", "work\_identifiers": ["The Band"], "artist\_or\_creator\_identifiers": ["The Band"], "categories": ["Director information", "Production information"]}}

</tool\_call>

\par\noindent\rule{\linewidth}{0.2pt}

<think>

The search results mainly show various types of directors and band-related information, but do not directly answer who the director of "The Band" is. I realize that my search parameters may not be precise enough. "The Band" may refer to a specific film or work, rather than the general concept of a band.

Let me reanalyze the problem: I need to find information about the director of the specific work "The Band." This should be a film or a specific musical work. I should search again, focusing on "The Band" as a specific film or work, rather than the general concept of a band.

</think>

<tool\_call>

{"name": "culture\_arts\_sports\_search", "arguments": {"query": "The Band film director information", "work\_identifiers": ["The Band"], "artist\_or\_creator\_identifiers": ["The Band"], "categories": ["Director information", "Production information"]}}

</tool\_call>

\par\noindent\rule{\linewidth}{0.2pt}

<think>

Great! Now I have found the key information. From the first search result, I can see:

"The Band (film) The Band, (also known as The Troupe) is an Israeli comedic musical, first shown in April 1978, about an army singing group in 1968. Three new members are hazed at first, one of them falls in love, and the choir's leader, in preparation for a television appearance, rehearses the troupe so much they revolt. The movie was initially unsuccessful but has gained in popularity. The choir's leader is based on real-life composer, arranger, and music director of the Nahal troupe, Yair Rosenblum. The film was directed by Avi Nesher."

This clearly shows that "The Band" is a film, and the director is Avi Nesher. The first search result already provides the complete answer.

</think>

<answer>Avi Nesher</answer>

 \\

\midrule 
\textbf{Final Answer:} Daniel Roher~\xmark & \textbf{Final Answer:} Avi Nesher~\cmark \\

\bottomrule
\end{tabular}
\end{table*}
\FloatBarrier

\subsection{Prompt Design}
\label{app:prompt}

\begin{tcolorbox}[colback=green!5!white,
colframe=green!60!blue,
width=1.08\textwidth,
left=1mm, right=1mm,
                    rounded corners,
                    coltitle=black, title=System Prompt]
                  
\# Role

You are a helpful assistant responsible for answering the user’s original question. You are adept at designing complex tool-calling sequences and multi-step reasoning chains. During the reasoning process, you must integrate a reflection mechanism to support self-monitoring, error correction, and dynamic optimization.

\# Task Description:

You have three response modes:

- Mode 1: Based on the user's original question, reason about a plan and select the appropriate tool(s) to call.

- Mode 2: Based on the result from a tool, analyze it and decide to proceed with another tool-calling.

- Mode 3: Based on the result from a tool, analyze it and provide the final answer directly.
Interaction Flow:

- Response to a user message: Analyze the original question, formulate a thought process, and call a tool.

- Response to a tool message: Interpret the tool's result and decide whether to continue with tool-calling or to output the final answer.

\# Output Structures

You must adhere to one of the following two output structures. Each output must be a direct response to the most recent user or tool message.

1.Structure 1 (Tool-Calling):

<think>Your thought process goes here.</think> + <tool\_call>The schema for the tool-calling goes here.</tool\_call>

2.Structure 2 (Providing a final answer):

<think>Your thought process goes here.</think> + <answer>Your final answer goes here.</answer>

\# Norms for the Thinking Part:

  - All outputs must begin with <think> and end with </think>. Keep the thinking process concise, ideally under 200 words. Inside <think>...</think>, you must perform the following steps:
  
  1.Identify the problem type:
  
  - Multi-hop question: State that the problem needs to be decomposed into sub-problems that will be solved through iterative tool-calling. The result of each step will inform the next sub-problem. Finally, all retrieved information will be aggregated to form the final answer.
  
  - Single-hop question: State that only a single round of tool-calling is necessary.
  
  2.Analyze the tool result (if applicable):
  
  - If you are calling another tool, explain how the previous result helps construct the new sub-problem.
  
  - If you are providing the final answer, explain that you have gathered sufficient information to conclude.
  3.Perform a reflection check across these dimensions:
  
  - Tool-Calling Consistency: Does the tool selection and parameterization mentioned in <think> match the actual invocation in <tool\_call>?
  
  - Logical Coherence: Do the reasoning steps within <think> logically support the subsequent tool-calling or the final answer?
  
  - Answer Consistency: Does the analysis within <think> align with the final answer provided in <answer>?
  
\# Norms for the Tool-Calling Part

Each tool-calling round must use one of the following formats:

1.Single tool-calling format:

<tool\_call>\textbackslash n...\textbackslash n</tool\_call>
2.Multiple tool-calling format:

Use consecutive <tool\_call>\textbackslash n...\textbackslash n</tool\_call> tags, separated by one newline character.
For example, when calling two tools sequentially:<tool\_call>\textbackslash n...\textbackslash n</tool\_call>\textbackslash n<tool\_call>\textbackslash n...\textbackslash n</tool\_call>  

\# Norms for the User Response Part

1.When providing the final answer, it must be enclosed in <answer>...</answer> tags.

2.The <answer>...</answer> tag must only contain the final answer itself. Do not include explanations, the reasoning process, or any other text.

\# Tools

You may call one or more functions to assist with the user query.

You are provided with function signatures within <tools></tools> XML tags:
<tools>...</tools>

For each function call, return a json object with function name and arguments within <tool\_call></tool\_call> XML tags:

<tool\_call>{"name": <function-name>, "arguments": <args-json-object>}</tool\_call>  
\end{tcolorbox}

\begin{tcolorbox}[colback=pink!20!white,
colframe=pink!60!white,
width=1.08\textwidth,
left=1mm, right=1mm,
rounded corners,
                  coltitle=black, title=User Prompt]
The original question from the user is: \{   \}
\end{tcolorbox}




\end{document}